\documentclass{article}
\usepackage{xcolor}

\usepackage{ijcai25}

\usepackage{times}
\usepackage{soul}
\usepackage{url}
\usepackage[hidelinks]{hyperref}
\usepackage[utf8]{inputenc}
\usepackage[small]{caption}
\usepackage{graphicx}
\usepackage{amsmath}
\usepackage{amsthm}
\usepackage{booktabs}
\usepackage{algorithm}
\usepackage{algorithmic}
\usepackage[switch]{lineno}

\title{VP Lab: a PEFT-Enabled Visual Prompting Laboratory for Semantic Segmentation}

\author{
Niccolo Avogaro$^1$$^,$$^2$\and 
Thomas Frick$^1$\and 
Yagmur G. Cinar$^1$\and 
Daniel Caraballo$^1$\and \\
Cezary Skura$^1$\and 
Filip M. Janicki$^1$\and 
Piotr Kluska$^1$ \and
Brown Ebouky$^1$$^,$$^2$\and \\
Nicola Farronato$^1$$^,$$^2$\and 
Florian Scheidegger$^1$\and 
Cristiano Malossi$^1$\and 
Konrad Schindler$^2$\and \\
Andrea Bartezzaghi$^1$\and  
Roy Assaf$^1$\and 
Mattia Rigotti$^1$\\
\affiliations
$^1$IBM Research, $^2$ETH Zürich\\
}

\begin{document}

\maketitle

\begin{abstract}
    Large-scale pretrained vision backbones have transformed computer vision by providing powerful feature extractors that enable various downstream tasks, including training-free approaches like visual prompting for semantic segmentation.
    Despite their success in generic scenarios, these models often fall short when applied to specialized technical domains where the visual features differ significantly from their training distribution. To bridge this gap, we introduce VP Lab, a comprehensive iterative framework that enhances visual prompting for robust segmentation model development.
    At the core of VP Lab lies E-PEFT, a novel ensemble of parameter-efficient fine-tuning techniques specifically designed to adapt our visual prompting pipeline to specific domains in a manner that is both parameter- and data-efficient.
    Our approach not only surpasses the state-of-the-art in parameter-efficient fine-tuning for the Segment Anything Model (SAM), but also facilitates an interactive, near-real-time loop, allowing users to observe progressively improving results as they experiment within the framework.
    By integrating E-PEFT with visual prompting, we demonstrate a remarkable 50\% increase in semantic segmentation mIoU performance across various technical datasets using only 5 validated images, establishing a new paradigm for fast, efficient, and interactive model deployment in new, challenging domains. This work comes in the form of a demonstration\footnote{Demonstration can be found in the \href{https://research.ibm.com/projects/visual-prompting}{\textcolor{blue}{\underline{official project page}.}}}
\end{abstract}

\section{Introduction}
Foundation Models have revolutionized machine learning by shifting from task-specific models to generalist ones pretrained on large, diverse datasets and fine-tuned for various downstream tasks \cite{bommasani2022opportunitiesrisksfoundationmodels}.
In computer vision, models like Segment Anything Model (SAM) \cite{kirillov2023segment}, CLIP \cite{radford2021learningtransferablevisualmodels}, and DINOv2 \cite{oquab2024dinov2learningrobustvisual} enable powerful functionalities such as semantic segmentation and classification in a zero-shot manner. Their versatile capabilities are central to many applications, including innovative approaches in applied computer vision, e.g.~for visual inspection~\cite{rigotti}.
This work focuses on semantic image segmentation, a key computer vision task essential for applications in medical imaging, autonomous driving, and visual inspection. We aim to develop a human-computer interaction workflow for few-shot open-world segmentation that efficiently addresses real-world, out-of-domain use cases.

We build on the existing visual prompting literature \cite{liu2024matchersegmentshotusing}, \cite{matcher_vp}, \cite{avogaro2025telleffectivelypromptingvisionlanguage}, which introduced pipelines relying on foundation models - in particular DINOv2 for feature matching and SAM to segment objects of interests based on minimal user annotations of reference images.
While these pipelines support accurate visual prompting in generic vision domains \cite{matcher_vp}, they often struggle with specialized technical applications.
Especially SAM, despite its broad capabilities, faces challenges in generating coherent segmentation masks for technical, domain-specific objects. 
This paper focuses on addressing these challenges, proposing a solution for scenarios where visual prompting alone is inadequate and lacks scalability.

To address this challenge, as our main innovation, we introduce E-PEFT, a scalable ensemble of parameter-efficient fine-tuning techniques for the SAM that ensures fast convergence and delivers state-of-the-art results for test-time training when integrated into a visual prompting pipeline. To fully leverage this technique in an interactive scenario, we introduce a complementary label correction workflow, which provides efficient mask refinement capabilities.
Together, these developments notably enhance performance in complex, real-world scenarios where visual prompting based on generic pretrained models alone is inadequate.
With the integration of E-PEFT into a base framework like~\cite{matcher_vp}, we enable an unparalleled level of adaptability to novel domains without significantly sacrificing the interactivity between the user and the model, allowing users to see progressively better results as they continue experimenting within the new framework.
This evolution transforms a visual prompting framework into a \textit{Visual Prompting Laboratory} (VP Lab), empowering users to explore and flexibly address new challenging semantic segmentation use cases.


\section{Related Work}

Parameter-efficient fine-tuning techniques have become essential for adapting large pretrained foundation models to specific tasks with minimal computational resources. One of the pioneering techniques introduced by \cite{houlsby2019parameterefficient} consists of adding trainable adapter layers inside the backbone, instead of fine-tuning the model parameters. The Low-Rank Adaptation (LoRA) framework introduced by~\cite{hu2021loralowrankadaptationlarge} achieves this by injecting low-rank trainable matrices into the attention layers of Transformer architectures, allowing for efficient fine-tuning without modifying the entire model. Building upon LoRA, QLoRA \cite{dettmers2023qloraefficientfinetuningquantized} combines model quantization with low-rank adaptation, enabling fine-tuning of large language models on limited hardware by reducing memory usage. 

Visual Prompt Tuning (VPT) \cite{jia2022visualprompttuning}, adapts pretrained vision models to new tasks by learning a set of visual prompts, i.e.~tokens which effectively condition the model without altering its original parameters. Additionally, the IA3 method \cite{liu2022fewshotparameterefficientfinetuningbetter} offers a parameter-efficient approach by learning task-specific adapters, further enhancing the adaptability of large models to diverse tasks.

Several methods have been proposed for fine-tuning SAM \cite{kirillov2023segment}, each targeting different aspects of image segmentation. Notably, HQ-SAM \cite{ke2023segmenthighquality} focuses on improving segmentation quality by prioritizing high-fidelity outputs. BOFT \cite{liu2024parameterefficientorthogonalfinetuningbutterfly} offers a parameter-efficient solution through orthogonal fine-tuning, optimizing segmentation performance while reducing computational costs and \cite{zhong2024convolutionmeetsloraparameter} proposes an hybrid solution adding low-rank convolution adapter layers to the SAM backbone.

\section{Visual Prompting Lab}

\begin{figure}[t!]
    \centering
    \includegraphics[width=0.49\textwidth]{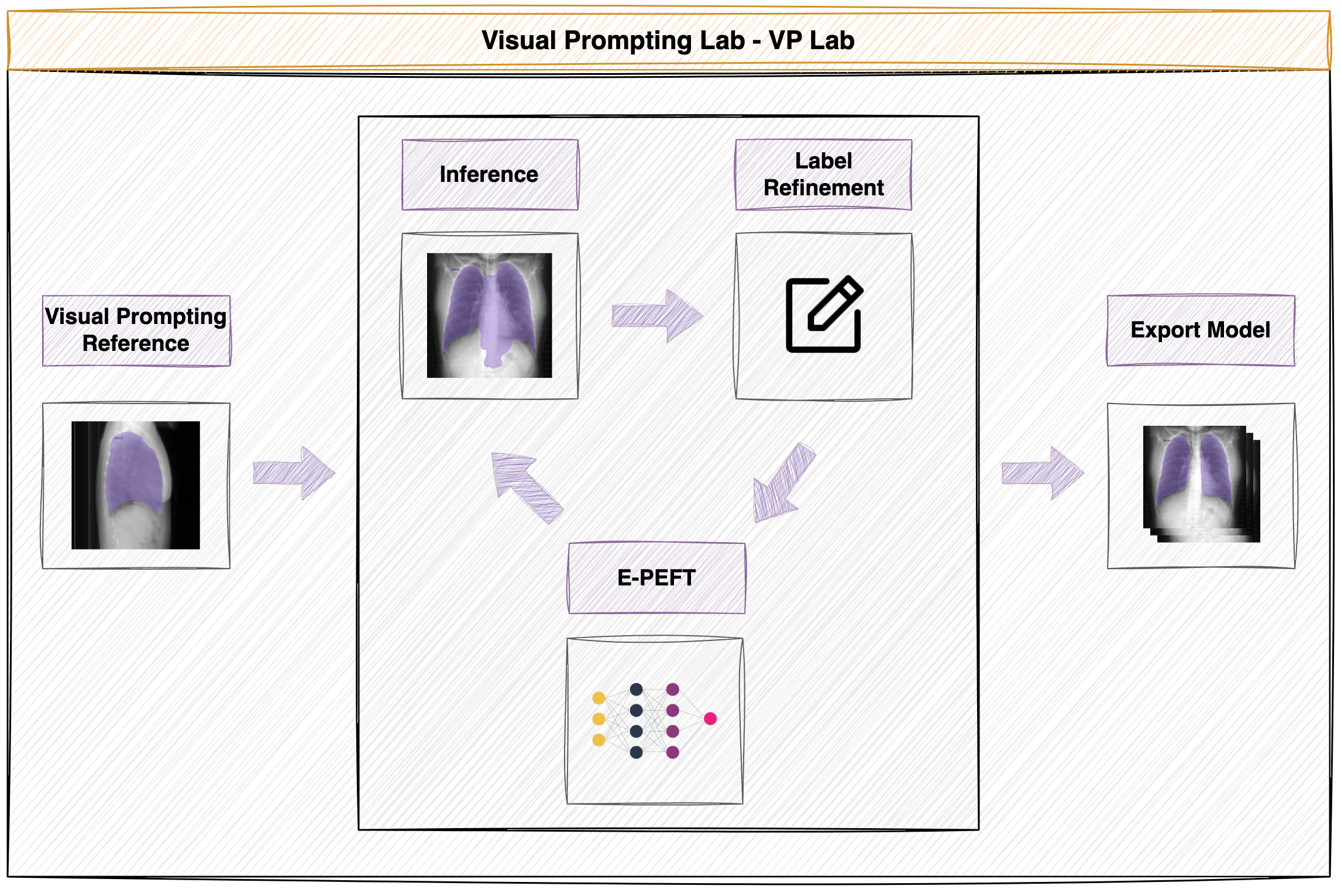} 
    \caption{VP Lab Workflow: Users prompt and validate a reference image, which guides predictions on target datasets. They can refine these predictions with a labeling tool, and feed them into a parameter-efficient fine-tuning process of the underlying model. After iterative improvements, the optimized model can be exported and deployed as needed.}
    \label{fig:vpl}
\end{figure}

\begin{table}[b!]
\centering
\caption{Performance of the proposed E-PEFT compared to the single PEFT baselines and to the SOTA HQ-SAM on Kvasir-Seg and HQ44k datasets.}
\begin{tabular}{lccc}
        \toprule
        Model & Params & Kvasir-Seg & HQ-44k \\
        \midrule
        SAM              & 0 & 72.88 & 84.49 \\
        Adapter        & 33.1K & 84.60 & 87.17 \\
        IA3            & 5.6K & 85.63 & 88.50 \\
        D-VPT          & 12.8K & 86.65 & 88.49 \\
        LORA           & 144.4K & 87.93 & 90.17 \\
        HQ-SAM         & 5.1M & 87.97 & 89.95 \\
        \midrule
        E-PEFT         & 201.1K & \textbf{88.97} & \textbf{90.50} \\
        \bottomrule
\end{tabular}

\label{tab:kvasir_hq44k_12epochs}
\end{table}

VP Lab (Figure \ref{fig:vpl}) is a scalable segmentation framework that enhances visual prompting to address complex technical use cases. The workflow consists of the following steps: i) \textbf{Visual prompting and validating a reference image} – the user provides points on the reference image to indicate the object class of interest. These input serve as a sparse prompt for SAM, which generates a reference mask; ii) \textbf{Generating pseudolabels} – the SoftMatcher \cite{matcher_vp} visual prompting algorithm computes pseudolabels for the target dataset; iii) \textbf{Refining labels} – labels are refined using an annotation tool in the web interface and validated by the human in the loop; iv) \textbf{Refining model} – the E-PEFT algorithm is used to fine-tune the SAM mask decoder in a matter of minutes, enhancing the segmentation capabilities of the full visual prompting pipeline and enabling it to handle new use cases not covered by the foundation model’s internet-scale training dataset. The refined model can be exported for production or used to further refine predictions by restarting from step ii). This loop can be repeated as needed.


The key innovations of our framework, compared to approaches like Softmatcher \cite{matcher_vp}, focus on creating a system capable of generating models for real-world technical use cases within minutes. Achieving this requires enabling users to make substantial changes to the model through efficient fine-tuning of the visual prompted foundation model. To promptly address challenging use cases, the fine-tuning process must rely on accurate labels, execute quickly, and consume minimal resources.
The first step for enabling this capability is the integration of a \textbf{label refinement tool} within the interface. This tool is essential for allowing users to contribute directly to the pipeline. By leveraging the interaction with the visual prompting framework, the labeling process becomes highly efficient, as users only need to refine existing outputs rather than annotate from scratch.
Then, the main enabler and innovation of this pipeline is an efficient, \textbf{parameter-efficient fine-tuning technique} designed for test-time adaptation. To support for rapid iterations between the model and user, the fine-tuning procedure must be fast, scalable, and resource-efficient. This key introduction serves as pivotal component driving the pipeline's strong performance on OOD tasks. The following section delves further into its details.

\textbf{Method.} Most existing approaches for fine-tuning SAM rely on complex pipelines that add millions of parameters to the models. These methods typically involve high-capacity tuning processes that require hours to complete and often require multiple GPUs.
To address this issue, we introduce \textbf{E-PEFT} (Ensemble of Parameter-Efficient Fine-Tuning), a scalable and efficient method optimized for the few-shot scenario. E-PEFT integrates multiple parameter-efficient techniques to enhance model adaptation when it comes to out-of-distribution tasks. The method targets SAM's decoder head, where parameter tuning has the greatest impact, enabling effective adaptation in low-data scenarios. E-PEFT combines LoRA, IA3, prompt tuning, and adapter modules, leveraging their orthogonal properties for improved performance.

\textbf{The motivation} behind E-PEFT stems from the fact that each method targets different parts of the model architecture, enabling seamless integration. This approach increases model capacity while maintaining a low parameter count, essential for fast convergence in low-data regimes. Specifically, LoRA decomposes the weight matrices of the decoder's linear layers into low-rank approximations. IA3 scales attention mechanisms and MLPs using three learned parameters per linear layer. Prompt tuning enhances learning by adding trainable memory tokens to the input sequence, which, in this case, are prepended to both of the decoder's input sequences. We also introduce a lightweight adapter module that shares the average image representation with the prompt through an MLP, in order for it to have a better initialization. E-PEFT leverages the complementary strengths of these techniques -- LoRA’s rapid convergence, IA3’s flexibility in scaling activations, and the capacity-enhancing effects of prompt tuning and the adapter module -- to optimize performance. The \textbf{synergy} between these methods allows E-PEFT to fine-tune only a small carefully selected subset of parameters that best maximize model performance on a given task. This is crucial for reducing computational costs and memory footprint, making it feasible to train on limited data or in resource-constrained environments.


\textbf{Experiments.} We validate our framework through experiments presented in Table \ref{tab:kvasir_hq44k_12epochs}. We highlight that E-PEFT outperforms the state-of-the-art HQ-SAM on the Kvasir-Seg \cite{jha2019kvasirseg} and HQ-44k \cite{ke2023segmenthighquality} datasets, notably with three orders of magnitude fewer parameters. These results further demonstrate that the PEFT techniques are orthogonal, and their combination results in better performance compared to using each technique individually.
\begin{table}[b!]
    \centering
    \caption{Semantic segmentation performance of E-PEFT with respect to HQ-SAM on Kvasir-Seg when considering 1- and 5-shot scenarios.}
    \begin{tabular}{@{}lcc@{}}
        \toprule
        Model & 1-shot & 5-shot \\
        \midrule
        HQ-SAM        & 59.30 & 75.71       \\
        E-PEFT    & \textbf{72.34}       & \textbf{82.25} \\
        \bottomrule
    \end{tabular}
    \label{tab:few_shot_res}
\end{table}
As summarized in Table~\ref{tab:few_shot_res}, our model excels in extremely limited data scenarios and shows a significant performance gap with respect to the state-of-the-art HQ-SAM, which fails to achieve good performance in 5-shot and 1-shot tuning, further demonstrating the effectiveness of our approach in low-data settings.

\begin{table}[h!]
\centering
\caption{Evaluation of the Softmatcher visual prompting method fine-tuned with E-PEFT across varying number of tuning shots. ``0-shot'' denotes the baseline without fine-tuning.}
\begin{tabular}{lcccc}
\toprule
Dataset             & 0-shot     & 5-shot    & 10-shot    & 40-shot    \\ 
\midrule
Kvasir-Inst. & 40.28         & 63.44         & 63.33         & 65.92         \\
PaxRay      & 36.39         & 48.61         & 51.19         & 50.97         \\
DeepCrack          & 11.96         & 19.27         & 21.64         & 23.71        \\
Corrosion CS      & \textcolor{white}{0}{4.06}         & \textcolor{white}{0}7.26         & \textcolor{white}{0}8.14         & \textcolor{white}{0}7.98         \\
\midrule
Average                   & \textbf{23.17} & \textbf{34.64} & \textbf{36.07} & \textbf{37.15} \\ 
\bottomrule
\end{tabular}

\label{tab:tuning_shots}
\end{table}

We evaluate the effectiveness of E-PEFT when integrated into an existing visual prompting pipeline (SoftMatcher) on four out-of-distribution datasets: Kvasir-Inst. \cite{kvasir}, PaxRay \cite{seibold2022detailedannotationschestxrays}, DeepCrack \cite{deepcrack}, and Corrosion CS \cite{Bianchi2021}, which cover technical engineering and medical use-cases. Results in Table~\ref{tab:tuning_shots} show that the visual prompting algorithm alone performs poorly in these technical domains. However, when integrated with E-PEFT, performance improves drastically. With just 5 images for training—\textbf{completed in under a minute on a NVIDIA V100 GPU}—average performance increases by 50\%, with some datasets showing up to an \textbf{80\% improvement}. As more images are added, performance continues to improve. Tuning with 40 images takes only 2.5 minutes, showcasing the method’s efficiency and effectiveness. This significant performance boost is made possible by the unique design choices in E-PEFT, which maximize both efficiency and capacity.

\textbf{Deployed service and frontend.} The interactive web interface is the core of VP Lab. It features a scalable architecture with an Angular-based~\cite{jain2014angularjs} frontend communicating via REST API with a Python backend, which organizes and submits compute tasks to PyTorch-based~\cite{pytorch} inference and training services, leveraging the available GPU resources.
When using the service, users can mark objects of interest on images using points, prompting the visual pipeline to generate precise segmentation masks for similar objects across target images. The initial results act as a support to iteratively build the final model. Users can manually refine masks using a set of annotation tools and use the resulting ground truth to fine-tune the model in a matter of minutes with extremely modest resources. This iterative process allows users to develop a deeper understanding of the model's behavior. By identifying its strengths and limitations, users can collaborate more effectively with the model, leading to improved outcomes. 

\textbf{Demonstration.} 
We present a fully interactive user-model workflow, showing its effectiveness in an example of real-world medical use case. The user begins by prompting the first image from a randomly sampled subset of the PaxRay dataset. This image then serves as input for the visual prompting pipeline, which returns predictions across the entire dataset. These predictions are not always sufficiently good, therefore the user refines them slightly using the annotation tool. The refined annotations are then used to tune the model, which in turn generates better results. The entire process takes less than a minute. Finally, a second round of label refinement and fine-tuning leads to virtually flawless results. 

\section*{Acknowledgments} 
This work is partially funded by the European Union’s Horizon Europe research and innovation program under grant agreement No.~101070408 (SustainML).


\bibliographystyle{named}
\bibliography{ijcai25}

\end{document}